\documentclass[sigconf]{aamas}

\usepackage{booktabs}          % three-line tables
\usepackage{array}             % >{...} column prefixes
\newcolumntype{L}[1]{>{\raggedright\arraybackslash}p{#1}}
\usepackage{amsmath}   % math (acmart loads amsmath already; harmless).
\usepackage{balance}           % balance last-page columns
\usepackage{graphicx}

\setcopyright{ifaamas}
\acmConference[AAMAS '27]{Proc.\\ of the 26th International Conference
on Autonomous Agents and Multiagent Systems (AAMAS 2027)}{May 3--7, 2027}
{Hanoi, Vietnam}{TBD (eds.)}
\copyrightyear{2027}
\acmYear{2027}
\acmDOI{}
\acmPrice{}
\acmISBN{}

\title[What to Admit and How to Present]{What to Admit and How to Present: Governing Persistent Memory in LLM Agents}

\author{Chang Liu}
\affiliation{
  \institution{Pazhou Laboratory}
  \city{Guangzhou}
  \country{China}
}
\author{Deliang Ding}

\begin{document}

\begin{abstract}
\looseness=-1 Persistent memory can improve personalization in LLM agents but can also induce sycophancy and cross-domain leakage. We distinguish two governance decisions: \emph{admission}, which determines what recalled information enters the working context, and \emph{presentation}, which determines how admitted information is expressed. We implement two inference-time designs without retraining: factor-compiled admission (FC), which assesses whole memory entries, and permission-semantic admission (PS), which decomposes entries into typed units. Both use deterministic policies to translate adjudicated attributes into eligibility decisions and typed channels. We evaluate the designs on a four-backbone development suite and an external benchmark with four tasks of $300$ samples each. Relative to verbatim injection, FC and PS reduce pooled judge-assessed failure rates on the external benchmark by $6.7$ and $8.8$ percentage points, respectively ($p=2.7\times10^{-7}$ and $p=4.1\times10^{-12}$). Development-set reductions in cross-domain leakage reach $29.5$ percentage points. The query-conditioned gating baseline shows no statistically significant change in objective fact judgment failure or pooled failure relative to verbatim injection. Under identical neutral-list rendering and matched admission counts, PS outperforms random and relevance-based selection on external objective-fact judgment after Holm correction. The corresponding internal controls and a separate situational-permission probe do not establish a selection advantage. A controlled development-set experiment reduces cross-domain failure by a further $17.5$ percentage points by tightening admission while holding presentation fixed ($p=1.6\times10^{-4}$). In contrast, no comparison between two renderings of identical adjudication outputs survives an exploratory Benjamini--Hochberg correction across nine task--metric pairs. Both designs increase personalization failures, and PS misses the preregistered improvement and personalization-preservation criteria. These results support evaluating admission and presentation separately: selection quality provides task-specific safety gains, while preserving beneficial memory use remains unresolved.
\end{abstract}


\maketitle

\section{Introduction}

\looseness=-1 Consider a memory entry stored by a conversational assistant: \textit{``The user has bought and self-custodied bitcoin since 2013.''} When the user asks \textit{``How do I buy bitcoin?''}, the memory provides a possible expertise signal that may justify a less introductory explanation. When the user asks \textit{``What should I invest in?''}, it provides background about the user's experience, but does not establish either a current investment preference or evidence that bitcoin is suitable for a new investment. For a query about resume writing, it is simply irrelevant. Correct behavior depends on two decisions that current memory-augmented agents do not separate: \emph{whether} the memory enters the working context at all, and \emph{how} whatever is admitted is organized and expressed in the prompt. Memories are typically injected verbatim, leaving the generator to resolve both questions implicitly---often defaulting to sycophantic validation or cross-domain leakage \citep{pulipaka2026persistbench}.

\looseness=-1 We call these two decisions the \textbf{admission} lever and the \textbf{presentation} lever. The distinction matters because the two levers have different costs, different failure modes, and---as this paper shows---very different effect sizes. Admission changes which information is available in the working context: excluded units cannot directly influence the response through memory injection, but useful information may also be withheld. Presentation preserves the admitted content while changing its organization and wording, although misleading information may still influence the response. Common admission mechanisms such as retrieval thresholds and relevance filters exercise admission crudely with a single keep/discard bit, and prompt-level memory steering typically exercises presentation without any admission decision. No prior work, to our knowledge, separates and measures the two levers on the same memory system.

\looseness=-1 This paper does so with a controlled decomposition. On the admission side we instantiate \emph{situated adjudication}: one batched adjudication call sees the query, the answer target, the task mode, and \emph{all} top-$k$ memories together (enabling cross-memory conflict detection), assigns attributes---never action decisions---and physically drops ineligible units. We evaluate two admission designs. \textbf{Factor-compiled admission} (FC) assesses each retrieved entry as a whole and uses deterministic code to map its attributes to typed channels. \textbf{Permission-semantic admission} (PS) first decomposes entries into typed units and maps their attributes to six permission categories: five permitted uses and \emph{deny}. Units assigned \emph{deny} are omitted from the generation prompt.
 On the presentation side we instantiate two sectioned renderings of the \emph{same} adjudicated verdicts---permission-note headers (PN) and information-type channel headers (CH)---grouping the same admitted units under differently labeled sections.

\looseness=-1 All confirmatory analyses were preregistered before data collection, with frozen prompts and assets. The evaluation runs on two surfaces: a four-backbone development suite spanning sycophancy, beneficial use, and cross-domain leakage ($n=122/91/200$ per dimension), and an external benchmark of four preference-memory tasks with $300$ samples each \citep{xiang2026memsyco}, judged end-to-end by a single external judge. Our findings:

\begin{itemize}\setlength\itemsep{1.5pt}\setlength\parsep{0pt}
\item \textbf{Adjudication-based pipelines cut judged harm; one controlled contrast isolates an admission move.} On the external benchmark, FC cuts pooled judged harm by $6.7$pp ($p=2.7\times10^{-7}$) and PS by $8.8$pp ($p=4.1\times10^{-12}$) against verbatim injection; they share adjudicated admission as well as situated analysis, typed judgment, and structured-use instructions, so we attribute the cuts to the adjudication-based pipelines as wholes rather than to admission alone (PS-vs-FC: pooled $p=.066$, not statistically significant). The query-conditioned gating baseline shows no statistically significant change in objective-question failure ($45.2\%\to45.5\%$, $p=1.0$) or pooled failure ($p=.08$). A separate controlled contrast---presentation, prompt, and judge fixed, only the injection set tightened---cuts cross-domain failure a further $17.5$pp ($p=1.6\times10^{-4}$) in one development configuration. Budget-matched controls, in which every comparison arm receives exactly as many memory units per query as PS, attribute part of the external gain to the selection itself (the internal study is preregistered; the external extension follows the same frozen analysis family and thresholds without a separate preregistration): at equal budget PS cuts objective-fact failure below both random-selection and relevance-ranking arms (Holm-corrected), while the same controls show no significant selection effect on the development surface. On the development suite, admission cuts cross-domain leakage by $29.5$pp on Qwen3.7-Max and $15.0$pp on Kimi-K2.6 (both surviving false-discovery correction).
\item \textbf{Between the two tested sectioned renderings, no robust difference.
} In a verdict-shared controlled experiment---same adjudication decisions, same admitted content, only the rendering differs---the single nominally significant cell ($+4.0$pp, $p=.043$) does not survive an exploratory BH correction across the nine task-metric pairs ($q{=}.39$), and a preregistered cross-backbone main test is $0/4$ significant. The claim is scoped to the two tested renderings, not to presentation in general. The only directly significant presentation effect we observe is a single-backbone design-phase observation ($9.9\%\to2.2\%$, two-sided $p=.039$), which we report as suggestive only.
\item \textbf{Failures under memory--evidence conflict persist under every tested governance wording.} Misleading rates remain $82.3$--$85.0\%$ in the FC, PS, and rendering arms ($85.7\%$ under the gate baseline), against $91.7\%$ under verbatim injection; a no-memory arm shows $0\%$ misleading. Among tested configurations, only those that remove memories from the context eliminate the effect; no arm that retains the conflicting memory, under any wording we test, brings misleading below $82\%$. We read this as a residual-failure observation, not an established mechanism (Section~\ref{sec:VI}).
\item \textbf{Boundaries.} Both admission designs \emph{increase} missed-personal\-ization failures on the personalization task (FC $+7.6$pp, PS $+5.3$pp, both significant)---an overshoot observed for both designs; GPT-OSS-120B does not improve on leakage under the tested configurations ($p=.12$; a generation-time configuration does, supplement~S5
); and a preregistered effect-size gate is not met (pooled improvement $8.8$pp vs.\ a $10$pp threshold; personalization accuracy $-6.0$pp vs.\ a $3$pp no-degradation line).
\end{itemize}

\looseness=-1 Within the evaluated configurations, the results support prioritizing admission decisions, while providing no robust basis for preferring one tested rendering over the other. Our contribution is an experimental framework that distinguishes pipeline-level effects, presentation effects under fixed verdicts, and the effect of tightening admission in a controlled configuration.

\section{Related Work}

\subsection{Memory-Augmented Agents and Their Risks}
\label{sec:II-A}

{\sloppy\looseness=-1 Generative Agents \citep{park2023generative} introduced the memory stream; MemGPT \citep{packer2023memgpt} adapted paging to long conversations; subsequent systems add forgetting curves \citep{zhong2024memorybank}, compression \citep{pan2025secom}, multi-layer stores \citep{wang2025mirix}, and managed extraction pipelines \citep{chhikara2025mem0,rasmussen2025zep}. PersistBench \citep{pulipaka2026persistbench} formalizes the risk side: memory-induced sycophancy (median $97.8\%$ failure across 18 frontier LLMs) and cross-domain leakage (median $53\%$), alongside beneficial personalization. MemSyco-Bench \citep{xiang2026memsyco} extends preference-memory risk to five controlled tasks including memory--evidence conflict and scope overreach; MIST \citep{bensal2026recalling} concurrently measures sycophancy in memory-augmented models. We use PersistBench's dimensions as a development surface and MemSyco-Bench's tasks as the external surface, and we study mitigation rather than measurement.\par}

\subsection{Defenses: Filtering, Steering, and Removal}
\label{sec:II-B}

\looseness=-1 Admission-type defenses exist at three stages of the memory lifecycle. Storage-side admission control \citep{zhang2026admission} scores candidate memories before they are written, optimizing memory-bank quality rather than what a retained memory may later do; retrieval-layer defenses \citep{lewis2020retrieval,srivastava2026causal} exercise admission as a binary relevance decision at query time; closest to us, MemGate \citep{zhang2026beyond} learns a query-conditioned neural gate over memory representations. Because sycophancy-inducing memories are topically relevant by construction, a natural concern is that similarity-space gates may not separate them from beneficial memories; we probe this empirically by reimplementing its gating policy on our substrate (Table~\ref{tab:ext}, Gate). Our context-side admission operates after retrieval and before generation, assessing whole entries in FC and typed units in PS. Presentation-layer structuring has also been tried as a standalone defense: partitioning memories into structured regions at inference reduces cross-domain leakage while leaving sycophancy largely unchanged \citep{hannoon2026structured}---consistent with our finding that presentation alone carries limited weight when the admitted content is unchanged. Instruction-level steering \citep{agrawal2026gepa} exercises presentation without admission: a standing prompt tells the generator how to treat memories, with effect sizes that vary widely across models. Deletion-centric approaches---certified removal \citep{bourtoule2021machine}, unlearning \citep{maini2024tofu}---remove stored content altogether, which is orthogonal to our question of how a \emph{retained} memory may act; in adversarial settings, state-contamination work \citep{wang2026state} shows that summarization can launder toxic content into classifier-clean memory and that post-hoc cleaning of stored summaries fails, consistent with our memory--evidence conflict finding even though their threat is adversarial toxicity rather than ordinary governance failure. Our adjudication differs from all of these: it is situated (decisions depend on the query, not just the memory) and separated from action---the adjudicator judges facts, deterministic code compiles them into structure, and FC operates on whole entries, whereas PS operates on typed units; excluded content is omitted from the generation context.

\begin{figure}[t]
\centering
\includegraphics[width=.92\columnwidth]{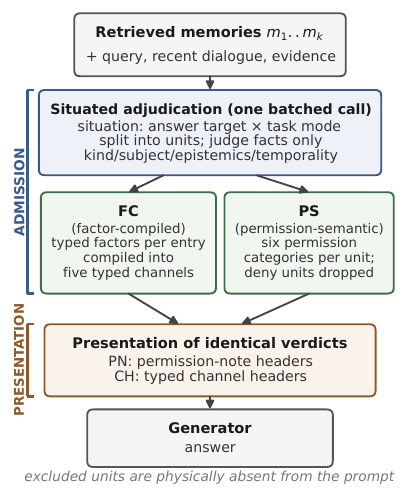}
\caption{Governance decomposed into two levers over one adjudication. \emph{Admission} (upper half): a single batched adjudication assigns attributes to retrieved memories in situ; two admission designs (FC, PS) compile the adjudicated attributes to decide what enters the context, and excluded material is physically absent. \emph{Presentation} (lower half): the same admitted verdicts are rendered under either sectioned form (PN vs.\ CH)
---the verdict-shared comparison of Section~\ref{sec:V-C} varies only this stage.}
\label{fig:pipeline}
\end{figure}

\section{Separating Admission from Presentation}
\label{sec:III}

\subsection{Problem Formulation}
\label{sec:III-A}

\looseness=-1 At generation time an agent retrieves the top-$k$ memories $M=\{m_1,\dots,m_k\}$ for query $q$. Governance composes two maps. \emph{Admission} $A$ maps each retrieved entry $m$ to typed units $U(m)$---with $U(m){=}\{m\}$ when the design judges whole entries, as in FC---and selects which of them reach the working context; \emph{presentation} $P$ decides the surface form of the prompt that encodes the admitted units. The deployed default injects everything retrieved verbatim. Both levers act at inference time---no retraining, no deletion, no edit of stored content (Figure~\ref{fig:pipeline}). Three complementary designs measure how far each lever moves outcomes: pipeline-level comparisons quantify adjudicated admission (plus compilation) against verbatim injection; a verdict-shared experiment isolates presentation wording under identical verdicts---a control condition specific to the rendering comparison, not to the FC-vs-PS comparison---and a single-variable contrast isolates tightening admission with presentation fixed (Section~\ref{sec:V-D}).

\subsection{Situated Adjudication (the Admission Lever)}
\label{sec:III-B}

\looseness=-1 One batched adjudication call receives the query, the recent dialogue, benchmark-provided evidence, and \emph{all} $k$ memories together, and returns a structured verdict. Its design has three fixed properties, shared across both admission designs and both evaluation surfaces:

\begin{enumerate}\setlength\itemsep{1pt}\setlength\parsep{0pt}

\item \textbf{Attribute assessment separated from policy decisions.} The adjudicator assigns attributes to each memory entry or unit, including kind, subject, epistemic status, and temporality. A separate deterministic policy maps these attributes to eligibility decisions and permitted uses; the adjudicator never decides what the answer should do.
\item \textbf{Situation awareness.} A situation block judges the \emph{answer target} (about the user, a third party, or the world) and the \emph{task mode} (factual, recommendation, planning, support) before any unit verdict, so the same memory can be eligible for one query and ineligible for another.
\item \textbf{Physical exclusion.} Units compiled out of the context are dropped from the prompt entirely---absence, not annotation of absence.
\end{enumerate}

\looseness=-1 Two designs instantiate the compilation step, both frozen before their respective confirmatory runs:

\textbf{FC: factor-compiled admission.} FC assesses each retrieved entry as a whole and assigns typed influence factors: scope, subject match, role, epistemic support, temporal status, and utility. For mixed entries, an \texttt{allowed\_content} field may restrict which portion can be used. Deterministic code maps these factors to one of five typed channels: constraints to follow, personalization signals, non-authoritative user views, supporting context, and uncertain context. The policy is asymmetric: constraints and current preferences fail open---when attributes leave their status uncertain, the item is retained in a supporting channel---whereas beliefs and unverified facts fail closed---uncertainty excludes them. Excluded memories never enter the generation context.

\textbf{PS: permission-semantic admission.} PS first decomposes each memory into typed units. For example, ``I prefer brand A because its battery lasts longest'' yields a preference unit and a factual-claim unit, allowing eligibility to differ within the same entry. The adjudicator assigns attributes to each unit, and deterministic code maps these attributes to one of six permission categories: five permitted uses, such as applying a constraint or mentioning a user view, plus \emph{deny}. Units assigned \emph{deny} are omitted from the generation prompt. PS is the design-phase iteration of the adjudication schema (unit splitting; attribute-to-permission compilation)
; its rendering uses the channel form below.

\subsection{Channel vs.\ Permission-Note Rendering (the Presentation Lever)}
\label{sec:III-C}

\looseness=-1 Holding adjudicated verdicts and admitted content fixed, we compare two renderings. \textbf{PN (permission-note rendering)}: admitted units are grouped under section headers, each stating what that section's information may be used for in this answer.
 \textbf{CH (channel)}: units are grouped under typed channel headers that state what each group may be used for (the variant that won a design-phase comparison on the development backbone; Section~\ref{sec:V-C}). Both are sectioned renderings encoding the same permission semantics over the same admitted units; they differ in section labeling and organization---permission-note labels vs.\ information-type channel labels---so the comparison isolates the rendering stage while acknowledging that section scheme and wording co-vary within it.

\subsection{Preregistration and Asset Freezing}
\label{sec:III-D}

\looseness=-1 Every confirmatory comparison in this paper was preregistered with frozen assets before its data collection: the cross-backbone suite (arm definitions, dimensions, test families), the external gate and three-way comparison (task set, pooling, thresholds), and the verdict-shared rendering experiment (including the verdict-sharing mechanism: the PN arm runs first and writes verdicts to a ledger; the CH arm reads the ledger without re-adjudication, guaranteeing identical admission decisions). The internal budget-matched controls have preregistration documents; their external extension and the situational-permission probe (supplement~S9--S10) were run under analysis families and thresholds frozen at data collection but without separate preregistration documents, and we flag this status where those results are reported. Adjudication prompts, channel headers, and permission schemas are held fixed from the development phase; no prompt was tuned after any external run began. An independent audit script recomputed every reported statistic from raw per-sample labels; missing judge scores ($2$ of $9{,}600$ external judgments) are handled by pairwise exclusion, never zero-filling.

\section{Experimental Setup}
\label{sec:IV}

\subsection{Two Evaluation Surfaces}
\label{sec:IV-A}

\textbf{Development surface: PersistBench} \citep{pulipaka2026persistbench}, three dimensions over pre-injected memory pools: sycophancy ($n{=}122$ after retrieval filtering), beneficial use ($n{=}91$), and cross-domain leakage ($n{=}200$); the official failure criterion (judge score $\geq 3$ on a 1--5 rubric) is adopted throughout. The development evaluation compares \textbf{Bare} (verbatim injection, no adjudication), \textbf{PS+PN}, and \textbf{PS+CH}.

\looseness=-1 \textbf{External surface: MemSyco-Bench} \citep{xiang2026memsyco}, four tasks $\times$ $300$ samples on a MemZero-style persistent-memory substrate \citep{chhikara2025mem0}: \emph{objective fact judgment} (ofj; does a stored user preference bend an objectively answerable question), \emph{contextual scope control} (csc; does a preference leak outside its scope), \emph{memory--evidence conflict} (mec; does a stored claim override provided evidence), and \emph{personalized memory use} (pm; is an appropriate stored preference actually applied). Each task's primary binary failure criterion is judged per sample: misconception adoption (ofj), out-of-scope preference use (csc), misled-by-conflicting-memory (mec), and preference-not-used (pm). External arms: \textbf{Bare}, \textbf{Gate} (query-conditioned gating \citep{zhang2026beyond}, reimplemented on the same substrate), \textbf{FC}, \textbf{PS}, \textbf{PS+PN}, \textbf{PS+CH}, and a \textbf{NoMem} ablation (same questions, no memory system).

\subsection{Models and Judge Protocol}
\label{sec:IV-B}

\looseness=-1 Development generators: Qwen3.7-Max, MiniMax-M3, Kimi-K2.6, and GPT-OSS-120B \citep{qwen2025qwen3,lai2026minimax,kimi2025k2,openai2025gptoss}; the external surface generates with DeepSeek-V4-Pro \citep{deepseek2024v3}. All generators run temperature $0$, thinking disabled, except Kimi-K2.6 (API-enforced $0.6$) and GPT-OSS-120B (reasoning effort \texttt{low}, the closest available approximation). The FC/PS adjudicator is DeepSeek-V4-Pro on both surfaces; the supplement's generation-time system (Two-Phase) runs its own phase-1 scorer, with per-run configuration documented in supplement~S5. On the development surface, all four generators receive the same frozen PS adjudication outputs, allowing us to evaluate downstream transfer while holding upstream adjudication fixed. A single external judge (GLM-5.2, temperature $0$, thinking disabled) scores every arm on every surface \citep{zheng2023judging}; the judge is never a generator, and all within-study comparisons share judge, threshold, and prompts. To check the judge, two human annotators independently labeled a stratified sample of $360$ development-surface outputs ($120$ per dimension) under the same score$\geq$3 failure rubric; all annotator--judge and annotator--annotator disagreements were resolved by item-level third-party adjudication, and the judge agrees with the resulting consensus labels at binary $\kappa{=}0.70$ ($n{=}352$; supplement~S3).
 No human validation was performed for the external task rubrics.

\subsection{Statistical Analysis}
\label{sec:IV-C}

\looseness=-1 All comparisons are paired at the sample level. Binary paired outcomes are analyzed using two-sided exact McNemar tests; internal comparisons against the five-seed random baseline use a paired permutation test on per-case seed means. Internal budget-matched studies use Holm correction over preregistered families, and the external budget-matched extension applies the same correction without a separate preregistration document, while the cross-backbone comparisons use Benjamini--Hochberg correction within their specified families. We report corrected $q$-values for the cross-backbone families and uncorrected raw $p$-values elsewhere as preregistered, marking nominal results explicitly. A pooled external test concatenates per-sample binary failures across the four task criteria ($n{=}1{,}199$; $1{,}200$ minus one missing judge score, excluded pairwise). An effect-size gate preregistered for PS requires pooled harm improvement $\geq10$pp at $p<.05$ \emph{and} no task accuracy drop $>3$pp.

\subsection{Preregistration, Audit, and Availability}
\label{sec:IV-D}

\looseness=-1 Three preregistration documents (cross-backbone suite; external gate and three-way comparison; verdict-shared rendering experiment) were frozen before their respective data collections and are released with the paper, alongside per-sample binary labels for every arm, the frozen adjudication and rendering assets, and an independent audit script that recomputes every statistic reported here from raw labels. Parse-failure rates are zero on both surfaces. The $9{,}600$ external judgments decompose as eight arms $\times$ four tasks $\times$ $300$ (Bare, FC, PS, PS+PN, PS+CH, NoMem, the Gate baseline of Table~\ref{tab:ext}, and the generation-time arm reported in the supplement); the two missing scores are excluded pairwise rather than zero-filled.

\section{Results}
\label{sec:V}

\subsection{Admission Transfers Across Backbones}
\label{sec:V-A}

\looseness=-1 Table~\ref{tab:dev} shows the development suite: adding adjudication to verbatim injection, under either rendering. The cross-domain leakage dimension carries the transfer: Qwen3.7-Max improves $44.0\%\to14.5\%$ (PN) / $18.0\%$ (CH) and Kimi-K2.6 $37.0\%\to22.5\%$ / $22.0\%$. The Qwen3.7-Max and Kimi-K2.6 cross-domain comparisons survive BH correction under both renderings ($q\approx0$ for Qwen; Kimi $q=.0034$ under PN and $q=.0011$ under CH): CH comparisons are evaluated within the preregistered family, and PN comparisons within a separate exploratory family. MiniMax-M3 improves on the same dimension at $9.5\%\to4.0\%$ but does not survive BH correction ($q=.053$), nor does the Qwen3.7-Max comparison on sycophancy under the CH family ($q=.070$); the Kimi-K2.6 sycophancy comparison survives under the exploratory PN family ($q{=}.019$) but narrowly misses the preregistered CH family ($q{=}.051$), and the Qwen3.7-Max one survives under the exploratory PN family ($q{=}.0034$). In a separate development-set comparison on DeepSeek-V4-Pro, a grouping-only con\-trol---sections labeled with benchmark-provided domain labels, no adjudication---does not explain these gains: relative to it the full method improves sycophancy ($32.0\%\to19.7\%$, $p=.006$) and cross-domain leakage ($19.0\%\to5.0\%$, $p{=}1.5{\times}10^{-5}$), while we detect no statistically significant improvement from grouping alone over verbatim injection (both $p{\geq}.25$; supplement~S2).
 Sycophancy improves directionally on all four backbones under CH yet no individual cell survives correction---we read this as a consistent direction without independent significance, not as four confirmations. No comparison shows a statistically significant increase in beneficial-use failure (all $p\ge.052$). These results do not establish equivalence or exclude smaller utility losses at this sample size.

\begin{table}[t]
\centering
\caption{Development surface: failure rate (\%) under verbatim injection (Bare) vs.\ PS+PN and PS+CH; paired exact McNemar, raw $p$. Dimensions: syc = sycophancy, ben = beneficial use, xd = cross-domain leakage. \textbf{Bold} = survives BH correction within its own family: CH cells against the preregistered cross-backbone family (CH vs.\ Bare on syc/xd, four backbones); PN cells against a separate exploratory family.}
\label{tab:dev}
{\footnotesize\setlength{\tabcolsep}{3pt}
\begin{tabular}{@{}llrrrrr@{}}
\toprule
Backbone & Dim & Bare & PS+PN & $p$ & PS+CH & $p$\\
\midrule
MiniMax-M3 & syc & 31.1 & 27.0 & .40 & 29.5 & .83\\
MiniMax-M3 & ben & 16.5 & 18.7 & .82 & 16.5 & 1.0\\
MiniMax-M3 & xd & 9.5 & 4.0 & .052 & 4.0 & .027\\
Qwen3.7-Max & syc & 50.0 & \textbf{36.1} & \textbf{.0009} & 41.8 & .053\\
Qwen3.7-Max & ben & 0.0 & 0.0 & 1.0 & 0.0 & 1.0\\
Qwen3.7-Max & xd & 44.0 & \textbf{14.5} & \textbf{${<}10^{-4}$} & \textbf{18.0} & \textbf{${<}10^{-4}$}\\
Kimi-K2.6 & syc & 43.4 & \textbf{32.0} & \textbf{.0094} & 32.8 & .019\\
Kimi-K2.6 & ben & 0.0 & 1.1 & 1.0 & 2.2 & .50\\
Kimi-K2.6 & xd & 37.0 & \textbf{22.5} & \textbf{.0013} & \textbf{22.0} & \textbf{.0003}\\
GPT-OSS-120B & syc & 45.9 & 38.5 & .064 & 37.7 & .041\\
GPT-OSS-120B & ben & 0.0 & 1.1 & 1.0 & 1.1 & 1.0\\
GPT-OSS-120B & xd & 64.0 & 70.0 & .12 & 69.0 & .20\\
\bottomrule
\end{tabular}}
\end{table}

\looseness=-1 The last row is the boundary: GPT-OSS-120B enters with $64\%$ leakage---the worst bare rate in the suite---and adjudication does not significantly reduce its leakage rate ($64.0\%\to70.0\%$, $p=.12$: no significant improvement, and not a significant harm). We therefore do not extrapolate a bare-leakage threshold from this single non-improving backbone; Section~\ref{sec:VI} returns to why.

\subsection{Admission Cuts Harm on the External Benchmark}
\label{sec:V-B}

\looseness=-1 Table~\ref{tab:ext} is the external three-way comparison. Against verbatim injection, both admission designs cut judged harm on every safety-oriented task and in the pooled test: FC $46.2\%\to39.5\%$ ($-6.7$pp, $p=2.7\times10^{-7}$) and PS $46.2\%\to37.4\%$ ($-8.8$pp, $p=4.1\times10^{-12}$). The task-matched objective-question cell is the largest single effect: $45.2\%\to20.4\%$ ($-24.7$pp, $p=7.6\times10^{-19}$), with answer accuracy rising $56.2\%\to75.9\%$. Because FC and PS share adjudicated admission as well as situated analysis, typed judgment, and structured-use instructions, we attribute these cuts to the adjudication-based pipelines as wholes; the single-variable contrast of Section~\ref{sec:V-D} isolates one admission move within such a pipeline, and PS-vs-FC is not statistically significant (pooled $p=.066$), so the comparison does not establish an advantage of either design. A budget-matched control on the same benchmark (every comparison arm receives exactly as many memory units per query as PS; internal study preregistered, external extension not separately preregistered; supplement~S10) shows the selection itself contributes to the external gain: at equal budget PS cuts objective-fact failure below both a random-selection arm ($23.3\%$ vs.\ $31.7\%$, Holm-adjusted $p{=}.001$) and a relevance-ranking arm ($44.3\%$, adjusted $p{<}.0001$), with accuracy rising in step and preference-use metrics showing no significant difference (at most one sample's difference across $300$ pairs); on the internal development surface the same controls show no statistically significant selection advantage.

\begin{table}[t]
\centering
\caption{External benchmark: judged failure rate (\%) by task, $n{=}300$ per cell (one ofj cell $299$ after pairwise exclusion). Paired exact McNemar vs.\ Bare. \textbf{Gate} = our reimplementation of the query-conditioned gating policy of MemGate~\citep{zhang2026beyond} on the same substrate (memories admitted by query-conditioned relevance, no adjudication). pm failure = appropriate stored preference not applied (lower is better). $^{\dagger}$ marks a failure-rate \emph{increase} relative to Bare (Section~\ref{sec:V-E}). Task abbreviations: ofj = objective fact judgment, csc = contextual scope control, mec = memory--evidence conflict, pm = personalized memory use.}
\label{tab:ext}
{\footnotesize\setlength{\tabcolsep}{3pt}
\resizebox{\columnwidth}{!}{%
\begin{tabular}{@{}lrrrrrrr@{}}
\toprule
Task & Bare & Gate & $p$ & FC & $p$ & PS & $p$\\
\midrule
ofj & 45.2 & 45.5 & 1.0 & 23.7 & $1.4{\times}10^{-13}$ & 20.4 & $7.6{\times}10^{-19}$\\
csc & 33.3 & 22.7 & .0004 & 27.0 & .042 & 25.3 & .0049\\
mec & 91.7 & 85.7 & .0003 & 85.0 & $8.8{\times}10^{-5}$ & 84.0 & $3.4{\times}10^{-5}$\\
pm & 14.7 & 22.7 & .0012$^{\dagger}$ & 22.3 & .0022$^{\dagger}$ & 20.0 & .0365$^{\dagger}$\\
\midrule
pooled ($n{=}1{,}199$) & 46.2 & 44.1 & .08 & 39.5 & $2.7{\times}10^{-7}$ & 37.4 & $4.1{\times}10^{-12}$\\
\bottomrule
\end{tabular}}}

\end{table}

\looseness=-1 The Gate column locates our increment against a direct retrieval-side baseline. Query-conditioned gating improves the two scope-conflict tasks (csc $-10.7$pp, mec $-6.0$pp, both $p<.001$) but shows no statistically significant change in objective-question failure ($45.2\%\allowbreak\to45.5\%$, $p=1.0$) and worsens personalization ($+8.0$pp, $p=.0012$); its pooled change is not significant ($p=.08$). Directly paired against this gate, PS improves the objective-fact task (McNemar $p{=}2.4{\times}10^{-17}$ over $300$ valid pairs) but shows no significant advantage on scope control ($p{=}.42$) or memory--evidence conflict ($p{=}.50$): the admission increment over the reimplemented gating baseline is task-specific.
 This is the failure mode our adjudication is designed to repair: preference-bearing beliefs are topically relevant to the query, and in this evaluation the reimplemented gate does not separate them on the objective-fact task, while situated adjudication separates on epistemics and scope relative to the query. PS reduces ofj failure by $24.7$pp relative to Bare. Its personalization failure increase is numerically smaller than Gate's ($+5.3$pp vs.\ $+8.0$pp), although this difference has not been separately tested.

PS has a lower observed failure rate than FC on all four tasks, but none of the direct comparisons is statistically significant. Pooled failure decreases from $39.5\%$ to $37.4\%$ ($p=.066$; discordant pairs 98 vs.\ 73). The task-level $p$-values are $.14$ for ofj, $.62$ for csc, $.70$ for mec, and $.36$ for pm. Personalization failure is also numerically lower under PS ($20.0\%$ vs.\ $22.3\%$; $p=.36$); these results do not establish PS's superiority over FC.

\subsection{Holding Verdicts Fixed, No Robust Difference Between Renderings}
\label{sec:V-C}

\looseness=-1 The verdict-shared experiment compares PS+PN vs.\ PS+CH with \emph{identical} adjudication decisions---the PN arm writes verdicts to a ledger that the CH arm reads without re-adjudication---so any difference is attributable to the rendering change (section labels and organization) alone.
 Table~\ref{tab:render} shows the result across four tasks and nine task-metric pairs. The single nominally significant cell (ofj accuracy $73.7\%\to77.7\%$, $p=.043$) fails BH correction across the nine task-metric pairs ($q=.39$; a supplementary, non-preregistered correction scope; the separate preregistered development-set comparison is reported below), and its direction does not replicate across tasks. The two pm rows are perfectly symmetric ($27/27$ and $28/28$ discordant pairs, both $p=1.0$). A preregistered cross-backbone main test of PN-vs-CH on the development surface (beneficial dimension, four backbones) is likewise $0/4$ significant (all $p\ge.50$): CH improves beneficial-use failure on one backbone, worsens it on one, and leaves the observed rate unchanged on two. Pooling all $1{,}200$ pairs across the four tasks, the PN--CH failure difference is $0.8$pp (exact McNemar $p{=}.43$) with a 95\% paired CI of $[-2.7, +1.0]$pp for CH$-$PN---a bound on the \emph{four-task-average} difference between these two renderings, not on wording effects in general and not per task (task-level effects of opposite sign could cancel; we saw none exceeding $+4.0$pp nominal).

\begin{table}[t]
\centering
\caption{Verdict-shared rendering experiment: identical admission decisions, two sectioned renderings (PN vs.\ CH).
 Paired exact McNemar; $n{=}300$ per row.}
\label{tab:render}
{\footnotesize\setlength{\tabcolsep}{4pt}
\begin{tabular}{@{}llrrr@{}}
\toprule
Task & Metric & PN & CH & $p$\\
\midrule
ofj & accuracy & .737 & .777 & .043\\
ofj & misconception & .220 & .190 & .12\\
csc & accuracy & .170 & .160 & .76\\
csc & scope violation & .253 & .237 & .60\\
mec & misled & .833 & .823 & .68\\
mec & accuracy & .003 & .007 & 1.0\\
pm & accuracy & .603 & .603 & 1.0\\
pm & use pass & .600 & .600 & 1.0\\
pm & preference used & .827 & .807 & .43\\
\bottomrule
\end{tabular}}
\end{table}

\looseness=-1 The only directly significant presentation effect in this paper is the design-phase comparison that selected CH: on the single development backbone, beneficial-use failure fell $9.9\%\to2.2\%$ (two-sided $p=.039$; one-sided as preregistered $p=.0195$). This is a single-backbone, uncorrected observation that did not generalize as an independent effect on any other backbone or on the external surface. We find no statistically robust difference between the two tested renderings at this sample size. This result applies to their combined differences in section labels and organization, and does not establish that presentation effects are absent in general.

\subsection{Memory Pressure, and Admission in Isolation}
\label{sec:V-D}

\looseness=-1 The mec task reveals a persistent failure under the tested governance configurations. On mec, misleading rates range from $82.3\%$ to $85.0\%$ across the FC, PS, and rendering arms, compared with $85.7\%$ under Gate and $91.7\%$ under Bare; in the rendering arms, objective accuracy is near zero ($.003$ and $.007$; Table~\ref{tab:render}, mec rows). The task is dominated by the conflicting premise: when it is present in the context, generators overwhelmingly follow it, and no wording of it that we test ($.833$ vs.\ $.823$ misled) repairs generation. With the memory system off, misleading drops to $0\%$---though this ablation removes the entire memory block (beneficial and conflicting content alike), so it establishes memory-dependence, not which content is responsible. The no-memory ablation sharpens the same point across tasks (Table~\ref{tab:nomem}): removing memory collapses personalization (pm accuracy $1.0\%$, preference use $2.0\%$) and zeroes out scope violations and misleading---memory is simultaneously the benefit and the hazard, and in our arms only interventions that change \emph{which} memory content reaches the context move both sides of that trade-off.

\looseness=-1\setlength{\emergencystretch}{2em} A single-variable experiment isolates that admission move directly. On the development surface's cross-domain dimension (DeepSeek-V4-Pro, $n{=}200$; supplement~S5), an annotated-memory arm---memories kept in the prompt with per-memory influence annotations, generation instructed to acknowledge but not adopt---fails $42.5\%$. Holding the same annotations, the same generation prompt, the same judge, and the same phase-1 scores fixed, and changing \emph{only} the injection set (memories whose maximum influence score falls below $0.5$ are excluded from the generation prompt), failure falls to $25.0\%$ (paired exact McNemar $p=1.6\times10^{-4}$, $59{:}24$ discordant pairs). Presentation identical, admission tightened: a further $17.5$pp. This controlled contrast shows that tightening admission reduces cross-domain failure in the evaluated configuration. Separately, the verdict-shared experiments find no statistically robust difference between the two tested renderings.

\begin{table}[t]
\centering
\caption{No-memory ablation compared with the PS+CH arm. Values are percentages. Removing memory sharply reduces personalization accuracy and preference use, while yielding zero observed scope violations and misleading responses; this ablation removes both beneficial and conflicting content (Section~\ref{sec:V-D}).}
\label{tab:nomem}
{\footnotesize\setlength{\tabcolsep}{3pt}
\begin{tabular}{@{}lrr@{}}
\toprule
Task & NoMem & PS+CH\\
\midrule
pm: accuracy / pref.\ used & 1.0 / 2.0 & 60.3 / 80.7\\
csc: accuracy / violation & 0.0 / 0.0 & 16.0 / 23.7\\
mec: misled & 0.0 & 82.3\\
\bottomrule
\end{tabular}}
\end{table}

\looseness=-1 On ofj, a no-memory arm run under our own protocol---same generator, judge, items, and answer prompts; the only intended difference is the absence of memory injection---scores $75.3\%$ accuracy ($226/300$). Accuracy is $56.2\%$ ($168/299$) under verbatim injection and $75.9\%$ ($227/299$) under PS. The latter two estimates use the $299$ valid paired observations, whereas the no-memory arm has valid scores for all $300$ items. Injecting preference-bearing memories is associated with a $19.1$pp accuracy drop before any governance, and adjudicated admission recovers it.

\subsection{Boundaries}
\label{sec:V-E}

\looseness=-1 \textbf{The personalization overshoot is shared by both designs.} Both admission pipelines \emph{increase} pm failures---appropriate preferences left unapplied---FC $14.7\%\to22.3\%$ ($p=.0022$) and PS $14.7\%\to20.0\%$ ($p=.0365$), with pm accuracy dropping $66.3\%\to57.0\%$ (FC) and $\to60.3\%$ (PS). Two structurally different designs overshooting on the same task is a shared observation; our design does not isolate whether the cost is inherent to the admission move or to shared pipeline components. \textbf{The preregistered gate is not met.} The PS gate demanded pooled improvement $\geq10$pp \emph{and} no task accuracy drop $>3$pp; measured: $8.8$pp ($1.2$pp below the threshold) and $-6.0$pp pm accuracy (beyond the permitted $3$pp decrease). The pooled improvement is statistically significant ($p{=}4.1{\times}10^{-12}$) and the ofj failure reduction is $24.7$pp; neither result changes the gate outcome. \textbf{One backbone does not improve under the tested configurations.} GPT-OSS-120B's elevated bare leakage ($64\%$) shows no statistically significant change under the tested admission-plus-rendering configurations, PS+PN and PS+CH ($64.0\%\to70.0\%/69.0\%$, $p=.12$/$.20$); a generation-time configuration (supplement~S5) cuts
 the same backbone's cross-domain failure materially, so the boundary is configuration-specific rather than a model limit (Section~\ref{sec:VI}). \textbf{One cross-batch gap.} Bare external runs precede governed runs by $\sim$30 hours (identical judge protocol and frozen assets, preregistered as run-once); we flag it. \textbf{The adjudicator under-triggers on fine-grained situational permissions.} A dedicated probe ($180$ paired queries spanning object-switch, context-switch, and opinion-use permissions; four arms; two backbones; supplement~S9) finds no paired-success gain of PS over an equal-budget relevance arm or bare injection (DeepSeek-V4-Pro $35.6\%$ vs.\ $32.8\%$/$31.1\%$, both $p\ge.22$), and memory injection is net-negative on Qwen3.7-Max ($7.2\%$ vs.\ $28.3\%$ without memory, $p{<}10^{-4}$). The adjudication logs suggest a possible source of failure: $56\%$ of restrict-worthy memory units are assigned CONTEXT, $8\%$ are denied, and only $11.9\%$ of restrict-bearing queries receive full governance; this observation does not isolate adjudicator errors from reference-label ambiguity or downstream generation failures. The side-level decomposition shows the governable content exists---memory presence moves success by up to $+53.8$pp on use-side queries and $-42$ to $-82$pp on restrict-side queries (supplement~S9)---while the probe does not isolate adjudicator errors from reference-label ambiguity or downstream generation failures.

\section{Discussion}
\label{sec:VI}

\looseness=-1 \textbf{Why the lever is at the context layer.} The mec observation generalizes: sycophancy-inducing and leaking memories are topically relevant by construction, so relevance-space filters may fail to separate them from beneficial ones without discarding both---consistent with the gate results of Section~\ref{sec:V-B}. Situated adjudication is designed to separate them on \emph{epistemics and scope relative to the query}---is this unit a verified fact or a user belief? is the answer about the user or the world?---dimensions orthogonal to topical similarity; on fine-grained situational switches, however, the situational-permission probe finds this separation does not transfer (supplement~S9). The mec results establish that misleading responses depend on the presence of the memory block under the tested conditions: misleading rates remain at $82.3$--$85.0\%$ in the FC, PS, and rendering arms, compared with $85.7\%$ under the gate baseline and $0\%$ under the no-memory arm; because the no-memory arm removes all recalled content, this comparison does not isolate the effect of selectively excluding conflicting units. This places the lever at the \emph{context} layer of the admission map (Section~\ref{sec:II-B})---after retrieval and before generation, decided per unit for the current query---where it complements rather than competes with storage-side write admission~\citep{zhang2026admission} and retrieval-side trustworthy search~\citep{zhang2026beyond}.

\looseness=-1 \textbf{Why GPT-OSS-120B does not improve under the tested configurations.} Its bare leakage ($64\%$) is the suite's worst, and the tested admission-plus-rendering configurations (PS+PN, PS+CH) show no statistically significant improvement ($64.0\%\to70.0\%/69.0\%$, $p=.12$/$.20$). In the separate precursor study, Two-Phase reduces this backbone's cross-domain failure from $84.0\%$ to $45.5\%$ (supplement~S5), indicating that the backbone can improve under another evaluated configuration; because the two studies use different configurations and baselines, this comparison does not identify the cause of the main evaluation's result. We leave the mechanism open.

\looseness=-1 \textbf{What presentation is still for.} Channel rendering remains our deployed default: it adds no model calls, it won the design-phase comparison on the development backbone (suggestively, single-backbone), and the verdict-shared experiment finds no statistically significant difference between CH and PN. The tested presentation strategies do not resolve memory--evidence conflict. This finding motivates further investigation of admission controls, but does not establish that other presentation strategies would be ineffective.

\looseness=-1 \textbf{Threats to validity.} (i) One judge scores all arms; judge validation is human-checked on the development rubric (binary $\kappa=0.70$ against two-annotator post-adjudication consensus labels, $n{=}352$; supplement~S3), not per external task rubric, though all comparisons share the judge, and the verdict-sharing design makes the rendering comparison insensitive to stable judge preferences (style-dependent judge bias is not ruled out). (ii) The pooled external test concatenates four task-level failure constructs (three safety-directional, one personalization-miss), i.e., an equal-weight average of task failure rates; pooling is preregistered, and because the pm construct moves \emph{against} admission the pooled estimate understates the safety-directional improvement (bootstrap details in supplement~S2). (iii) Cross-backbone breadth lives on the development surface; the external surface uses one generator. (iv) Reasoning was disabled where supported, and GPT-OSS-120B used the lowest available reasoning setting; the interaction of FC/PS with extended reasoning was not evaluated---an exploratory thinking-ON pilot for the precursor system appears in supplement~S6.

\section{Conclusion}
\label{sec:VII}

{\sloppy\setlength{\emergencystretch}{1.5em}\looseness=-1 We distinguish admission from presentation in the governance of persistent memory. On the external benchmark, FC and PS reduce pooled judged failure by $6.7$pp and $8.8$pp, respectively, while the query-conditioned gating baseline shows no statistically significant pooled improvement. A separate controlled experiment reduces cross-domain failure by a further $17.5$pp by tightening admission while holding presentation fixed. Pipeline-level reductions in cross-domain leakage also occur across multiple backbones. Budget-matched controls attribute part of the external objective-fact improvement to selection quality itself, while the same controls detect no significant selection effect on the internal suite. In contrast, no comparison between the two tested renderings survives multiple-comparison correction. Substantial limitations remain: misleading rates on memory--evidence conflict remain at $82.3$--$85.0\%$ in the FC, PS, and rendering arms, compared with $91.7\%$ under Bare; both admission pipelines increase personalization failures; PS misses the preregistered gate; and the tested configurations do not improve leakage on GPT-OSS-120B. These findings support evaluating admission and presentation separately and reporting safety gains alongside losses in beneficial memory use.\par}

\balance
\bibliographystyle{ACM-Reference-Format}
\bibliography{references}

%%% -*-BibTeX-*-
%%% Do NOT edit. File created by BibTeX with style
%%% ACM-Reference-Format-Journals [18-Jan-2012].

\begin{thebibliography}{25}

%%% ====================================================================
%%% NOTE TO THE USER: you can override these defaults by providing
%%% customized versions of any of these macros before the \bibliography
%%% command.  Each of them MUST provide its own final punctuation,
%%% except for \shownote{}, \showDOI{}, and \showURL{}.  The latter two
%%% do not use final punctuation, in order to avoid confusing it with
%%% the Web address.
%%%
%%% To suppress output of a particular field, define its macro to expand
%%% to an empty string, or better, \unskip, like this:
%%%
%%% \newcommand{\showDOI}[1]{\unskip}   % LaTeX syntax
%%%
%%% \def \showDOI #1{\unskip}           % plain TeX syntax
%%%
%%% ====================================================================

\ifx \showCODEN    \undefined \def \showCODEN     #1{\unskip}     \fi
\ifx \showDOI      \undefined \def \showDOI       #1{#1}\fi
\ifx \showISBNx    \undefined \def \showISBNx     #1{\unskip}     \fi
\ifx \showISBNxiii \undefined \def \showISBNxiii  #1{\unskip}     \fi
\ifx \showISSN     \undefined \def \showISSN      #1{\unskip}     \fi
\ifx \showLCCN     \undefined \def \showLCCN      #1{\unskip}     \fi
\ifx \shownote     \undefined \def \shownote      #1{#1}          \fi
\ifx \showarticletitle \undefined \def \showarticletitle #1{#1}   \fi
\ifx \showURL      \undefined \def \showURL       {\relax}        \fi
% The following commands are used for tagged output and should be
% invisible to TeX
\providecommand\bibfield[2]{#2}
\providecommand\bibinfo[2]{#2}
\providecommand\natexlab[1]{#1}
\providecommand\showeprint[2][]{arXiv:#2}

\bibitem[\protect\citeauthoryear{Agrawal, Tan, Soylu, Ziems, Khare, Opsahl-Ong,
  Singhvi, Shandilya, et~al\mbox{.}}{Agrawal et~al\mbox{.}}{2026}]%
        {agrawal2026gepa}
\bibfield{author}{\bibinfo{person}{Lakshya~A. Agrawal},
  \bibinfo{person}{Shangyin Tan}, \bibinfo{person}{Dilara Soylu},
  \bibinfo{person}{Noah Ziems}, \bibinfo{person}{Rishi Khare},
  \bibinfo{person}{Krista Opsahl-Ong}, \bibinfo{person}{Arnav Singhvi},
  \bibinfo{person}{Herumb Shandilya}, {et~al\mbox{.}}}
  \bibinfo{year}{2026}\natexlab{}.
\newblock \showarticletitle{{GEPA}: Reflective Prompt Evolution Can Outperform
  Reinforcement Learning}. In \bibinfo{booktitle}{\emph{Proceedings of the 14th
  International Conference on Learning Representations ({ICLR})}}.
\newblock


\bibitem[\protect\citeauthoryear{Bensal, Magnuson, Balagopalan, and
  Bikel}{Bensal et~al\mbox{.}}{2026}]%
        {bensal2026recalling}
\bibfield{author}{\bibinfo{person}{Shelly Bensal}, \bibinfo{person}{Axel
  Magnuson}, \bibinfo{person}{Aparna Balagopalan}, {and}
  \bibinfo{person}{Daniel~M. Bikel}.} \bibinfo{year}{2026}\natexlab{}.
\newblock \showarticletitle{Recalling Too Well: Sycophancy Evaluation and
  Mitigation in Memory-Augmented Models}.
\newblock \bibinfo{journal}{\emph{arXiv preprint arXiv:2606.10949}}
  (\bibinfo{year}{2026}).
\newblock


\bibitem[\protect\citeauthoryear{Bourtoule, Chandrasekaran, Choquette-Choo,
  Jia, Travers, Zhang, Lie, and Papernot}{Bourtoule et~al\mbox{.}}{2021}]%
        {bourtoule2021machine}
\bibfield{author}{\bibinfo{person}{Lucas Bourtoule}, \bibinfo{person}{Varun
  Chandrasekaran}, \bibinfo{person}{Christopher Choquette-Choo},
  \bibinfo{person}{Hengrui Jia}, \bibinfo{person}{Adelin Travers},
  \bibinfo{person}{Baiwu Zhang}, \bibinfo{person}{David Lie}, {and}
  \bibinfo{person}{Nicolas Papernot}.} \bibinfo{year}{2021}\natexlab{}.
\newblock \showarticletitle{Machine Unlearning}. In
  \bibinfo{booktitle}{\emph{Proceedings of the 42nd {IEEE} Symposium on
  Security and Privacy (S\&P)}}.
\newblock


\bibitem[\protect\citeauthoryear{Chhikara, Khant, Aryan, Singh, and
  Yadav}{Chhikara et~al\mbox{.}}{2025}]%
        {chhikara2025mem0}
\bibfield{author}{\bibinfo{person}{Prateek Chhikara}, \bibinfo{person}{Dev
  Khant}, \bibinfo{person}{Saket Aryan}, \bibinfo{person}{Taranjeet Singh},
  {and} \bibinfo{person}{Deshraj Yadav}.} \bibinfo{year}{2025}\natexlab{}.
\newblock \showarticletitle{{Mem0}: Building Production-Ready {AI} Agents with
  Scalable Long-Term Memory}.
\newblock \bibinfo{journal}{\emph{arXiv preprint arXiv:2504.19413}}
  (\bibinfo{year}{2025}).
\newblock


\bibitem[\protect\citeauthoryear{{DeepSeek-AI}}{{DeepSeek-AI}}{2024}]%
        {deepseek2024v3}
\bibfield{author}{\bibinfo{person}{{DeepSeek-AI}}.}
  \bibinfo{year}{2024}\natexlab{}.
\newblock \showarticletitle{{DeepSeek-V3} Technical Report}.
\newblock \bibinfo{journal}{\emph{arXiv preprint arXiv:2412.19437}}
  (\bibinfo{year}{2024}).
\newblock


\bibitem[\protect\citeauthoryear{Hannoon, Zhao, Narayan, Goyal, and
  Sheth}{Hannoon et~al\mbox{.}}{2026}]%
        {hannoon2026structured}
\bibfield{author}{\bibinfo{person}{Hakeem Hannoon}, \bibinfo{person}{Andrew
  Zhao}, \bibinfo{person}{Mihir Narayan}, \bibinfo{person}{Sharvin Goyal},
  {and} \bibinfo{person}{Ivaxi Sheth}.} \bibinfo{year}{2026}\natexlab{}.
\newblock \showarticletitle{Mitigating Over-Personalization in {LLMs} via
  Structured Memory}.
\newblock \bibinfo{journal}{\emph{arXiv preprint arXiv:2608.08300}}
  (\bibinfo{year}{2026}).
\newblock


\bibitem[\protect\citeauthoryear{{Kimi Team}}{{Kimi Team}}{2025}]%
        {kimi2025k2}
\bibfield{author}{\bibinfo{person}{{Kimi Team}}.}
  \bibinfo{year}{2025}\natexlab{}.
\newblock \showarticletitle{{Kimi K2}: Open Agentic Intelligence}.
\newblock \bibinfo{journal}{\emph{arXiv preprint arXiv:2507.20534}}
  (\bibinfo{year}{2025}).
\newblock


\bibitem[\protect\citeauthoryear{Lai, Xu, Yang, et~al\mbox{.}}{Lai
  et~al\mbox{.}}{2026}]%
        {lai2026minimax}
\bibfield{author}{\bibinfo{person}{Xunhao Lai}, \bibinfo{person}{Weiqi Xu},
  \bibinfo{person}{Yufeng Yang}, {et~al\mbox{.}}}
  \bibinfo{year}{2026}\natexlab{}.
\newblock \showarticletitle{{MiniMax} Sparse Attention}.
\newblock \bibinfo{journal}{\emph{arXiv preprint arXiv:2606.13392}}
  (\bibinfo{year}{2026}).
\newblock


\bibitem[\protect\citeauthoryear{Lewis, Perez, Piktus, et~al\mbox{.}}{Lewis
  et~al\mbox{.}}{2020}]%
        {lewis2020retrieval}
\bibfield{author}{\bibinfo{person}{Patrick Lewis}, \bibinfo{person}{Ethan
  Perez}, \bibinfo{person}{Aleksandra Piktus}, {et~al\mbox{.}}}
  \bibinfo{year}{2020}\natexlab{}.
\newblock \showarticletitle{Retrieval-Augmented Generation for
  Knowledge-Intensive {NLP} Tasks}. In \bibinfo{booktitle}{\emph{Advances in
  Neural Information Processing Systems 33 ({NeurIPS})}}.
\newblock


\bibitem[\protect\citeauthoryear{Maini, Feng, Schwarzschild, Lipton, and
  Kolter}{Maini et~al\mbox{.}}{2024}]%
        {maini2024tofu}
\bibfield{author}{\bibinfo{person}{Pratyush Maini}, \bibinfo{person}{Zhili
  Feng}, \bibinfo{person}{Avi Schwarzschild}, \bibinfo{person}{Zachary~C.
  Lipton}, {and} \bibinfo{person}{J.~Zico Kolter}.}
  \bibinfo{year}{2024}\natexlab{}.
\newblock \showarticletitle{{TOFU}: A Task of Fictitious Unlearning for
  {LLMs}}.
\newblock \bibinfo{journal}{\emph{arXiv preprint arXiv:2401.06121}}
  (\bibinfo{year}{2024}).
\newblock


\bibitem[\protect\citeauthoryear{{OpenAI}}{{OpenAI}}{2025}]%
        {openai2025gptoss}
\bibfield{author}{\bibinfo{person}{{OpenAI}}.} \bibinfo{year}{2025}\natexlab{}.
\newblock \bibinfo{title}{gpt-oss-120b}.
\newblock \bibinfo{howpublished}{Hugging Face:
  https://huggingface.co/openai/gpt-oss-120b}.
\newblock
\newblock
\shownote{Open-weights release.}


\bibitem[\protect\citeauthoryear{Packer, Wooders, Lin, et~al\mbox{.}}{Packer
  et~al\mbox{.}}{2023}]%
        {packer2023memgpt}
\bibfield{author}{\bibinfo{person}{Charles Packer}, \bibinfo{person}{Sarah
  Wooders}, \bibinfo{person}{Kevin Lin}, {et~al\mbox{.}}}
  \bibinfo{year}{2023}\natexlab{}.
\newblock \showarticletitle{{MemGPT}: Towards {LLMs} as Operating Systems}.
\newblock \bibinfo{journal}{\emph{arXiv preprint arXiv:2310.08560}}
  (\bibinfo{year}{2023}).
\newblock


\bibitem[\protect\citeauthoryear{Pan, Wu, Jiang, Luo, Cheng, Li, Yang, Lin,
  Zhao, Qiu, and Gao}{Pan et~al\mbox{.}}{2025}]%
        {pan2025secom}
\bibfield{author}{\bibinfo{person}{Zhuoshi Pan}, \bibinfo{person}{Qianhui Wu},
  \bibinfo{person}{Huiqiang Jiang}, \bibinfo{person}{Xufang Luo},
  \bibinfo{person}{Hao Cheng}, \bibinfo{person}{Dongsheng Li},
  \bibinfo{person}{Yuqing Yang}, \bibinfo{person}{Chin-Yew Lin},
  \bibinfo{person}{H.~Vicky Zhao}, \bibinfo{person}{Lili Qiu}, {and}
  \bibinfo{person}{Jianfeng Gao}.} \bibinfo{year}{2025}\natexlab{}.
\newblock \showarticletitle{{SeCom}: On Memory Construction and Retrieval for
  Personalized Conversational Agents}. In \bibinfo{booktitle}{\emph{The
  Thirteenth International Conference on Learning Representations ({ICLR})}}.
\newblock


\bibitem[\protect\citeauthoryear{Park, O'Brien, Cai, et~al\mbox{.}}{Park
  et~al\mbox{.}}{2023}]%
        {park2023generative}
\bibfield{author}{\bibinfo{person}{Joon~Sung Park}, \bibinfo{person}{Joseph~C.
  O'Brien}, \bibinfo{person}{Carrie~J. Cai}, {et~al\mbox{.}}}
  \bibinfo{year}{2023}\natexlab{}.
\newblock \showarticletitle{Generative Agents: Interactive Simulacra of Human
  Behavior}. In \bibinfo{booktitle}{\emph{Proceedings of the 36th Annual ACM
  Symposium on User Interface Software and Technology ({UIST})}}.
\newblock


\bibitem[\protect\citeauthoryear{Pulipaka, Chen, Sharma, Bajwa, Raina, and
  Sheth}{Pulipaka et~al\mbox{.}}{2026}]%
        {pulipaka2026persistbench}
\bibfield{author}{\bibinfo{person}{Sidharth Pulipaka}, \bibinfo{person}{Oliver
  Chen}, \bibinfo{person}{Manas Sharma}, \bibinfo{person}{Taaha~S. Bajwa},
  \bibinfo{person}{Vyas Raina}, {and} \bibinfo{person}{Ivaxi Sheth}.}
  \bibinfo{year}{2026}\natexlab{}.
\newblock \showarticletitle{{PersistBench}: When Should Long-Term Memories Be
  Forgotten by {LLMs}?}. In \bibinfo{booktitle}{\emph{Proceedings of the 43rd
  International Conference on Machine Learning ({ICML})}}.
\newblock


\bibitem[\protect\citeauthoryear{{Qwen Team}}{{Qwen Team}}{2025}]%
        {qwen2025qwen3}
\bibfield{author}{\bibinfo{person}{{Qwen Team}}.}
  \bibinfo{year}{2025}\natexlab{}.
\newblock \showarticletitle{{Qwen3} Technical Report}.
\newblock \bibinfo{journal}{\emph{arXiv preprint arXiv:2505.09388}}
  (\bibinfo{year}{2025}).
\newblock


\bibitem[\protect\citeauthoryear{Rasmussen, Paliychuk, Beauvais, Ryan, and
  Chalef}{Rasmussen et~al\mbox{.}}{2025}]%
        {rasmussen2025zep}
\bibfield{author}{\bibinfo{person}{Preston Rasmussen}, \bibinfo{person}{Pavlo
  Paliychuk}, \bibinfo{person}{Travis Beauvais}, \bibinfo{person}{Jack Ryan},
  {and} \bibinfo{person}{Daniel Chalef}.} \bibinfo{year}{2025}\natexlab{}.
\newblock \showarticletitle{{Zep}: A Temporal Knowledge Graph Architecture for
  Agent Memory}.
\newblock \bibinfo{journal}{\emph{arXiv preprint arXiv:2501.13956}}
  (\bibinfo{year}{2025}).
\newblock


\bibitem[\protect\citeauthoryear{Srivastava}{Srivastava}{2026}]%
        {srivastava2026causal}
\bibfield{author}{\bibinfo{person}{Saksham~Sahai Srivastava}.}
  \bibinfo{year}{2026}\natexlab{}.
\newblock \showarticletitle{Causal Intervention-Based Memory Selection for
  Long-Horizon {LLM} Agents}.
\newblock \bibinfo{journal}{\emph{arXiv preprint arXiv:2605.17641}}
  (\bibinfo{year}{2026}).
\newblock


\bibitem[\protect\citeauthoryear{Wang and Chen}{Wang and Chen}{2025}]%
        {wang2025mirix}
\bibfield{author}{\bibinfo{person}{Yu Wang} {and} \bibinfo{person}{Xi Chen}.}
  \bibinfo{year}{2025}\natexlab{}.
\newblock \bibinfo{title}{{MIRIX}: Multi-Agent Memory System for {LLM-Based}
  Agents}.
\newblock \bibinfo{howpublished}{arXiv preprint arXiv:2507.07957}.
\newblock


\bibitem[\protect\citeauthoryear{Wang, Goyal, Chen, and Sundaram}{Wang
  et~al\mbox{.}}{2026}]%
        {wang2026state}
\bibfield{author}{\bibinfo{person}{Yian Wang}, \bibinfo{person}{Agam Goyal},
  \bibinfo{person}{Yuen Chen}, {and} \bibinfo{person}{Hari Sundaram}.}
  \bibinfo{year}{2026}\natexlab{}.
\newblock \showarticletitle{State Contamination in Memory-Augmented {LLM}
  Agents}.
\newblock \bibinfo{journal}{\emph{arXiv preprint arXiv:2605.16746}}
  (\bibinfo{year}{2026}).
\newblock


\bibitem[\protect\citeauthoryear{Xiang, Chen, Tang, Wei, Ning, Lin, Zhang, and
  Su}{Xiang et~al\mbox{.}}{2026}]%
        {xiang2026memsyco}
\bibfield{author}{\bibinfo{person}{Zhishang Xiang}, \bibinfo{person}{Zerui
  Chen}, \bibinfo{person}{Yunbo Tang}, \bibinfo{person}{Zhimin Wei},
  \bibinfo{person}{Ruqin Ning}, \bibinfo{person}{Yujie Lin},
  \bibinfo{person}{Qinggang Zhang}, {and} \bibinfo{person}{Jinsong Su}.}
  \bibinfo{year}{2026}\natexlab{}.
\newblock \showarticletitle{MemSyco-Bench: Benchmarking Sycophancy in Agent
  Memory}.
\newblock \bibinfo{journal}{\emph{arXiv preprint arXiv:2607.01071}}
  (\bibinfo{year}{2026}).
\newblock


\bibitem[\protect\citeauthoryear{Zhang, Jiang, Wang, Behr, Zhao, Friedman, Chu,
  and Anoun}{Zhang et~al\mbox{.}}{2026b}]%
        {zhang2026admission}
\bibfield{author}{\bibinfo{person}{Guilin Zhang}, \bibinfo{person}{Wei Jiang},
  \bibinfo{person}{Xiejiashan Wang}, \bibinfo{person}{Aisha Behr},
  \bibinfo{person}{Kai Zhao}, \bibinfo{person}{Jeffrey Friedman},
  \bibinfo{person}{Xu Chu}, {and} \bibinfo{person}{Amine Anoun}.}
  \bibinfo{year}{2026}\natexlab{b}.
\newblock \showarticletitle{Adaptive Memory Admission Control for {LLM}
  Agents}.
\newblock \bibinfo{journal}{\emph{arXiv preprint arXiv:2603.04549}}
  (\bibinfo{year}{2026}).
\newblock


\bibitem[\protect\citeauthoryear{Zhang, Chen, Ma, Hu, He, Zhang, Liu, Yang,
  Zhang, and Jia}{Zhang et~al\mbox{.}}{2026a}]%
        {zhang2026beyond}
\bibfield{author}{\bibinfo{person}{Jiawen Zhang}, \bibinfo{person}{Kejia Chen},
  \bibinfo{person}{Jiachen Ma}, \bibinfo{person}{Yangfan Hu},
  \bibinfo{person}{Lipeng He}, \bibinfo{person}{Yechao Zhang},
  \bibinfo{person}{Jian Liu}, \bibinfo{person}{Xiaohu Yang},
  \bibinfo{person}{Tianwei Zhang}, {and} \bibinfo{person}{Ruoxi Jia}.}
  \bibinfo{year}{2026}\natexlab{a}.
\newblock \showarticletitle{Beyond Similarity: Trustworthy Memory Search for
  Personal {AI} Agents}.
\newblock \bibinfo{journal}{\emph{arXiv preprint arXiv:2606.06054}}
  (\bibinfo{year}{2026}).
\newblock


\bibitem[\protect\citeauthoryear{Zheng, Chiang, Sheng, Zhuang, Wu, Zhuang, Lin,
  Li, Li, Xing, Zhang, Gonzalez, and Stoica}{Zheng et~al\mbox{.}}{2023}]%
        {zheng2023judging}
\bibfield{author}{\bibinfo{person}{Lianmin Zheng}, \bibinfo{person}{Wei-Lin
  Chiang}, \bibinfo{person}{Ying Sheng}, \bibinfo{person}{Siyuan Zhuang},
  \bibinfo{person}{Zhanghao Wu}, \bibinfo{person}{Yonghao Zhuang},
  \bibinfo{person}{Zi Lin}, \bibinfo{person}{Zhuohan Li},
  \bibinfo{person}{Dacheng Li}, \bibinfo{person}{Eric~P. Xing},
  \bibinfo{person}{Hao Zhang}, \bibinfo{person}{Joseph~E. Gonzalez}, {and}
  \bibinfo{person}{Ion Stoica}.} \bibinfo{year}{2023}\natexlab{}.
\newblock \showarticletitle{Judging {LLM}-as-a-Judge with {MT-Bench} and
  {Chatbot Arena}}. In \bibinfo{booktitle}{\emph{Advances in Neural Information
  Processing Systems 36 ({NeurIPS} 2023), Datasets and Benchmarks Track}}.
\newblock


\bibitem[\protect\citeauthoryear{Zhong, Guo, Gao, Ye, and Wang}{Zhong
  et~al\mbox{.}}{2024}]%
        {zhong2024memorybank}
\bibfield{author}{\bibinfo{person}{Wanjun Zhong}, \bibinfo{person}{Lianghong
  Guo}, \bibinfo{person}{Qiqi Gao}, \bibinfo{person}{He Ye}, {and}
  \bibinfo{person}{Yanlin Wang}.} \bibinfo{year}{2024}\natexlab{}.
\newblock \showarticletitle{{MemoryBank}: Enhancing Large Language Models with
  Long-Term Memory}. In \bibinfo{booktitle}{\emph{Proceedings of the 38th
  {AAAI} Conference on Artificial Intelligence}}.
\newblock


\end{thebibliography}

\end{document}